\documentclass[conference]{IEEEtran}

\usepackage{eqnarray,amssymb,amsmath,amsthm,mathtools,amsfonts}
\usepackage{mathrsfs}
\usepackage[utf8]{inputenc}
\usepackage[T1]{fontenc}
\usepackage{graphicx}
\usepackage{subfigure}
\usepackage{color}
\usepackage{epstopdf}
\usepackage{cases}
\usepackage{cite}
\usepackage{bbm}
\usepackage{algorithm}
\usepackage{algpseudocode}
\usepackage{soul}
\newtheorem{theorem}{Theorem}


\begin{document}

\title{Federated Learning with Correlated Data: Taming the Tail for Age-Optimal Industrial IoT}

\author{\IEEEauthorblockN{Chen-Feng Liu and Mehdi Bennis}\IEEEauthorblockA{Centre for Wireless Communications, University of Oulu, Finland\\E-mail: \{chen-feng.liu, mehdi.bennis\}@oulu.fi}}

\maketitle

\begin{abstract}
While information delivery in industrial Internet of things demands reliability and latency guarantees, the freshness of the controller's available information, measured by the age of information (AoI), is paramount for high-performing industrial automation. The problem in this work is cast as a sensor's transmit power minimization subject to the peak-AoI requirement and a probabilistic constraint on queuing latency. We further characterize the \emph{tail behavior} of the latency by a generalized Pareto distribution (GPD) for solving the power allocation problem through Lyapunov optimization. As each sensor utilizes its own data to locally train the GPD model, we incorporate \emph{federated learning} and propose a local-model selection approach which accounts for correlation among the sensor's training data. Numerical results show the tradeoff between the transmit power, peak AoI, and delay's tail distribution. Furthermore, we verify the superiority of the proposed correlation-aware approach for selecting the local models in federated learning over an existing baseline.
\end{abstract}


\section{Introduction}
Delivering the monitored status data with ultra-reliable low-latency communication (URLLC) and having up-to-date information at the central controller (in control systems) are pivotal in industrial Internet-of-things (IoT) networks \cite{5GACIA19,ZhaImrPanCheLi19,VitZunSau19}. In this regard, the age of information (AoI) \cite{KauGruRaiKen11}, which is the elapsed time since the data was generated till the current time instant, has been considered as the information freshness measure for resource allocation and scheduling in industrial IoT settings \cite{WanCheLiPanVuc19,LiuHuaGu19,LiCheHuaGua20,LiuBen19,HsuLiuSamWeiBen20}.

\subsection{Related Work}
By assuming that the sensors update their status information over unreliable links, the work \cite{WanCheLiPanVuc19} focused on average AoI minimization subject to the sensors' transmit power constraints. Therein, a transmission scheduling policy was proposed.
The authors in \cite{LiuHuaGu19} studied the channel allocation problem in software-defined industrial IoT and aimed to minimize the maximal average AoI over the network.
Considering that the status data is transmitted via device-to-device (D2D) communication in an industrial wireless network, Li \emph{et al.}~\cite{LiCheHuaGua20} proposed a belief-based Bayesian reinforcement learning framework in which 
 D2D users optimize their dynamic channel and power allocation policies in a distributed manner. The objective in \cite{LiCheHuaGua20} was to maximize energy efficiency subject to AoI constraints.
Moreover, a centralized \cite{LiuBen19} and a distributed \cite{HsuLiuSamWeiBen20} dynamic power allocation policy for sensors were proposed in our prior works by taking into account the statistics of the maximal AoI over time and the AoI threshold violation probability, respectively. In \cite{LiuBen19}, we further investigated URLLC with respect to the information decoding error incurred by the finite blocklength transmission. 
Note that the end-to-end delay, including the transmission delay, queuing delay, and so forth, are incorporated in the AoI-based formulation \cite{KauGruRaiKen11}. In other words, when we allocate communication resources, the AoI performance are entangled with the delays. Furthermore, analyzing the \emph{tail behavior} of the delay distribution is one key enabler for URLLC \cite{BenDebPoo18}. However, while the aforementioned works provided interesting results, little attention has been paid to the joint investigation of the AoI performance and the delay's tail distribution in state-of-the-art industrial IoT. 
Although AoI threshold deviation can be related to the data queue length in vehicular communication \cite{AbdSumLiuBenSaa20} in which we aimed to reduce the excess AoI/queue length, we still lacked the joint investigation of the AoI and delay.

\subsection{Our Contribution}
In this work, focusing on the uplink of an industrial IoT network with multiple sensors, we study the {\bf power minimization problem which accounts for the peak AoI requirement and the tail distribution of the queuing delay}. Specifically, a URLLC constraint in terms of the threshold violation probability is imposed on the queuing delay whose analytic tail distribution formula is needed for allocating the sensor's transmit power via Lyapunov optimization. 
To address this, we invoke \emph{extreme value theory}, by which the tail behavior can be characterized by a generalized Pareto distribution (GPD), and incorporate \emph{federated learning} (FL) \cite{MahMooRamHamArc17} in order to alleviate the sensors' overheads of finding the characteristic parameters of the GPD. 
The outcome of FL is affected by the correlation among the sensor's empirical data for training the GPD model. However, in most FL-aided wireless communication systems, the training data are independent \cite{ElbSol20,elbir2020federated,WanCheYinSaaHonCuiPoo21}, or the correlation among the training data is neglected \cite{HsuLiuSamWeiBen20}. Instead, we take correlation among the training data into consideration and propose a correlation-aware approach for selecting the sensors' local models in FL.
We investigate the tradeoff between the average power consumption, peak AoI, and queuing delay's tail distribution by simulations. Regarding GPD-model training, the proposed model selection approach achieves a lower variance compared with the correlation-agnostic baseline.

\section{System Model}\label{Sec: System}

\begin{figure}[t]
	\begin{center}
	\includegraphics[width = \columnwidth]{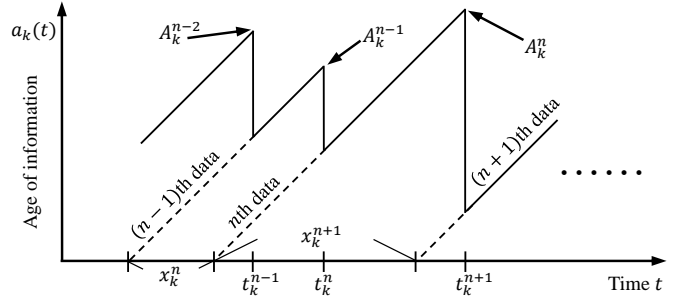}
	\caption{Communication timeline and AoI function of sensor $k$.}
	\label{Fig: AoI}
	\end{center}
\end{figure}

Consider the industrial IoT network composed of a set $\mathcal{K}$ of $K$ wireless sensors and a central controller.
The sensors monitor the factory environments and send the status data to the controller.
We assume that the sensors' data-sampling operations are triggered by random events. After sampling, the sensor transmits the status data immediately if the previous samples were uploaded. Otherwise, it queues in the data buffer for transmission. Let the sensor's sequentially sampled data be indexed by $n\in\mathbb{Z}^{+}$. Then we denote the queuing time of the $n$th data of sensor $k\in\mathcal{K}$ as $q_k^n\geq 0$. 
The total bandwidth $W$ is orthogonally and equally allocated to all sensors.
Given that the sensor $k$ allocates transmit power $P_k^n$ in its $n$th transmission, the corresponding transmission time is
\begin{equation}
T_k^n = \frac{KD}{W\log_2\big(1+\frac{Kh_k^nP_k^n}{WN_0}\big)}
\end{equation}
 with data size $D$. Here, $h_k^n$ is the channel gain, including path loss and channel fading, between sensor $k$ and the controller in the $n$th transmission, and $N_0$ is the power spectral density of the additive white Gaussian noise. Fig.~\ref{Fig: AoI} shows the communication timeline and AoI function of sensor $k$. Therein, $t_k^n$ is the time instant at which the controller receives the $n$th data. We denote the AoI as $a_k(t)$ which is the function of time index $t\in\mathbb{R}^{+}$ and measured at the controller. At time instant $t_k^n$, the age of the controller's newly received information, i.e., the $n$th data, is $q_k^n+ T_k^n$. Then the information age increases linearly with time. Hence, the AoI function can be mathematically defined as
\begin{equation}
a_k(t)=q_k^n+ T_k^n+t-t_k^n,~\forall\,t\in[t_k^n,t_k^{n+1}),n\in\mathbb{Z}^{+}.
\end{equation}
When the $n$th data is completely delivered to the controller, we have the peak AoI of the $(n-1)$th data (i.e., lifetime of the previous data)  as
\begin{align}
A_k^{n-1}&=\lim\limits_{\tau \to 0^+}a_k(t_k^{n}-\tau)\notag
\\&=a_k(t_k^{n-1})+\max\{ x_k^{n}-a_k(t_k^{n-1}),0\}+T_k^{n},
\end{align}
where $x_k^{n}>0$ represents the inter-arrival time between the $(n-1)$th data and $n$th data. Additionally, we can straightforwardly find the mathematical expression of the queuing time of sensor $k$'s $(n+1)$th data as
\begin{equation}
q_k^{n+1}=\max\{q_k^n+ T_k^n- x_k^{n+1},0\}.
\end{equation}
Further note that $x_k^{n+1}$ and $q_k^{n+1}$ may be unknown when we allocate transmit power $P_k^n$. Finally, for each sensor $k\in\mathcal{K}$, inter-arrival time $x_k^{n+1},\forall\,n\in\mathbb{Z}^{+}$, is identically distributed and can be correlated.\footnote{We assume positive correlation in this work.} The statistics of data arrivals are identical and independent among all sensors. One applicable scenario is that various sensors separately monitor the temperatures of the identical manufacturing processes in different factories.

\section{Peak AoI and URLLC-Aware Power Allocation}

\subsection{Problem Formulation}
Due to the continuous changes of the factory environment status, the controller's available information becomes outdated as time elapses. The aged information may further deteriorate the control system performance.
In order to suppress this deficiency, we consider a cost function $f_k^n = \frac{1}{\beta}(A_k^{n-1})^{\beta}$   for the peak AoI and impose a long-term time-averaged constraint $\lim\limits_{N \to \infty} \frac{1}{N}\sum^N_{n=1}\mathbb{E}[ f_k^n] \leq f_{\rm th},\forall\,k\in\mathcal{K}$, with a predetermined parameter $\beta \geq 1$ and the cost threshold $f_{\rm th}$.
Regarding the URLLC requirement, we impose a probabilistic constraint on the queuing delay in each transmission $n\in\mathbb{Z}^{+}$ as $\Pr\{q_k^{n+1}> q_{\rm th}|q_k^{n},h_k^{n}\}\leq \epsilon $, where $q_{\rm th}$ and  $\epsilon$ are the delay threshold and tolerable threshold violation probability, respectively. Note that the concerned probability $\epsilon$ is very small.
For the purpose of prolonging the battery-limited sensor's lifetime, we study a power minimization problem
\begin{subequations}\label{origin}
\begin{IEEEeqnarray}{cl}
\underset{P_k^n}{\mbox{minimize}}
&~~ \lim\limits_{N \to \infty} \frac{1}{N} \sum^N_{n=1} P_k^n\label{main_problem}
\\\mbox{subject to}&~~\lim\limits_{N \to \infty} \frac{1}{N}\sum^N_{n=1}\mathbb{E}[ f_k^n] \leq f_{\rm th},\label{AoI_loss_bound} 
\\&~~\Pr\{q_k^{n+1}> q_{\rm th}|q_k^{n},h_k^{n}\}\leq \epsilon,~\forall\, n\in\mathbb{Z}^{+},\label{outg}
\\&~~0\leq P_k^n \leq P_{\max},~ \forall \,n \in \mathbb{Z}^+, \label{PW_max}
\end{IEEEeqnarray}
\end{subequations}
for each sensor $k\in\mathcal{K}$, in which $P_{\max}$ is the sensor's power budget. 
Here, the expectation in \eqref{AoI_loss_bound} is taken with respect to the stochastic wireless channel and inter-arrival time, whereas the conditional probability in \eqref{outg} is measured with respect to the randomness of inter-arrival time.
We further note that a closed-form expression of constraint \eqref{outg} in terms of $P_k^n$ is required for proceeding with problem \eqref{origin}. To address this demand, let us first rewrite \eqref{outg} as
\begin{multline} 
\Pr\{q_k^{n+1}> q_{\rm th}|q_k^{n},h_k^{n}\}=\Pr\{q_k^n+ T_k^n- x_k^{n+1}> q_{\rm th}\}
\\=\Pr\{X > -\ln(q_k^n+ T_k^n-q_{\rm th})  \} \leq \epsilon\label{out transfer 1}
\end{multline}
given $q_{\rm th}> 0$, where $X=-\ln (x_k^{n+1}),\forall\,n\in\mathbb{Z}^{+},k\in\mathcal{K}$. 
In other words, the full distribution of inter-arrival time gives the desired closed-form expression of \eqref{outg}, but the distribution function of any arbitrary random variable $X$ is not always available. Since we are concerned about the tail distribution of $X$ owing to the very small probability $\epsilon$, we can resort to the Pickands–Balkema–de Haan theorem which asymptotically characterizes the tail behaviors of general probability distributions \cite{Coles2001}.
\begin{theorem}[{\bf Pickands–Balkema–de Haan theorem}]\label{Thm: GPD}
Given a random variable $X$ with the complementary cumulative distribution function (CCDF) $\bar{F}_X(x)$ and a threshold $x_0$, as $x_0 \to \bar{F}^{-1}_X(0)$, the conditional CCDF of the excess value $Y|_{X>x_0}=X-x_0>0$ can be approximated by a GPD, i.e.,  $\bar{F}_{Y|X>x_0}(y)=\Pr(X-x_0> y|X>x_0) \approx (1+\xi y/\sigma)^{-1/\xi}$, with a scale parameter $\sigma>0$ and a shape parameter $\xi\in\mathbb{R}$.
 \end{theorem}
Thus, we consider a threshold $x_0<-\ln\big(q_k^n+ T_k^n-q_{\rm th}\big)$ and rewrite \eqref{out transfer 1} as
\begin{equation} 
\Pr\{X > -\ln(q_k^n+ T_k^n-q_{\rm th})|X>x_0  \}\leq \frac{\epsilon}{\bar{F}_X(x_0)}. \label{out transfer 2}
\end{equation}
Then given $\epsilon< \bar{F}_X(x_0) \ll 1 $, \eqref{out transfer 2} is equivalent to the minimal transmit power requirement
\begin{multline}
P_k^n \geq P_{k,\min}^{n}=\frac{WN_0}{Kh_k^n}\bigg[-1+\exp\bigg(\frac{KD\ln 2}{W}
\\\times\frac{1}{ q_{\rm th}-q_k^n+ \exp\big\{\frac{\sigma}{\xi}\big[1-\big(\frac{\epsilon}{\bar{F}_X(x_0)}\big)^{-\xi}\big]-x_0\big\}} \bigg)\bigg]\label{out transfer 3} 
\end{multline}
by applying the results in Theorem \ref{Thm: GPD} to \eqref{out transfer 2}. The characteristic parameters $\boldsymbol{\theta}\equiv(\sigma,\xi)$ of the GPD in \eqref{out transfer 3} can be estimated by statistical methods while $\bar{F}_X(x_0) $ is obtained empirically. We will elaborate the approach to find $\boldsymbol{\theta}$ in Section \ref{Sec: non-iid FL}.
Given a specific value of  $\boldsymbol{\theta}$, the power allocation problem \eqref{origin} in which we replace \eqref{outg} with \eqref{out transfer 3} is subsequently solved by using Lyapunov optimization \cite{Lyapunov}.

\subsection{Sensor's Transmit Power Allocation}
Let us first introduce a virtual queue $Z_k^n$ with the queue length evolution
\begin{equation}\label{Eq: Virtual queue}
Z_k^{n+1}=\max\{Z_k^n+f_k^n-f_{\rm th},0\}
\end{equation}
for the time-averaged constraint \eqref{AoI_loss_bound}. In this regard, we need to stabilize the virtual queue, i.e., $\lim\limits_{n\to\infty}\frac{\mathbb{E}[\lvert Z_k^n\rvert]}{n}= 0$, in order to ensure constraint \eqref{AoI_loss_bound}.
Then we derive an upper bound on the conditional Lyapunov drift-plus-penalty \cite{Lyapunov} by applying $(\max\{x,0\})^2\leq x^2$ to \eqref{Eq: Virtual queue}, i.e.,
\begin{multline}\label{Eq: Bound}
 \mathbb{E}\Big[\frac{1}{2}(Z_{k}^{n+1})^2-\frac{1}{2}(Z_{k}^{n})^2+VP_k^n
\Big|Z_k^n\Big]
\\ \leq \mathbb{E}\Big[\frac{1}{2}(Z_k^n+f_k^n-f_{\rm th})^2-\frac{1}{2}(Z_{k}^{n})^2+VP_k^n
\Big|Z_k^n\Big]
\\ \leq\frac{1}{2}(f_{\rm th})^2+ \mathbb{E}\Big[Z_k^n f_k^n+\frac{1}{2}(f_k^n)^2+VP_k^n
\Big|Z_k^n\Big].
\end{multline}
To jointly stabilize the virtual queue and optimize the sensor's transmit power, we aim to minimize the upper bound \eqref{Eq: Bound}  \cite{Lyapunov}.
To this goal, the sensor $k$ solves
 \begin{equation}\label{Eq: Per slot problem}
 \underset{P^n_{k,\min}\leq P_k^n\leq P_{\max}}{\mbox{minimize}}
 ~~\frac{Z_k^n \big(c_k^{n} + T_k^n\big)^{\beta}}{\beta} +\frac{\big(c_k^{n} + T_k^n\big)^{2\beta}}{2\beta^2} 
+ VP_k^n
 \end{equation}
in each transmission $n$ with the constant $c_k^{n}=a_k(t_k^{n-1}) +\max\{ x_k^{n}-a_k(t_k^{n-1}),0\}$.
Here, $V\geq 0$ is a parameter trading off AoI reduction and the optimality of power consumption.
Note that the convexity of problem \eqref{Eq: Per slot problem} can be straightforwardly verified. Thus, via differentiation, we obtain the sensor's transmit power in the $n$th transmission as $P_k^{n*}=\max\{\min\{\tilde{P}_k^n,P_{\max}\},P^n_{k,\min}\}$ in which $\tilde{P}_k^n$ satisfies
\begin{multline}
V=\frac{K^2Dh_k^n \ln 2}{W\Big[\ln\Big(1+\frac{Kh_k^n\tilde{P}_k^n}{WN_0}\Big)\Big]^2\big(WN_0+Kh_k^n\tilde{P}_k^n\big)}
\\\times \Bigg[Z_k^n \Bigg(c_k^{n} + \frac{KD}{W\log_2\Big(1+\frac{Kh_k^n\tilde{P}_k^n}{WN_0}\Big)}\Bigg)^{\beta-1}
\\+\frac{1}{\beta}\Bigg(c_k^{n} + \frac{KD}{W\log_2\Big(1+\frac{Kh_k^n\tilde{P}_k^n}{WN_0}\Big)}\Bigg)^{2\beta-1}\Bigg].
\end{multline}
After sending the status data, sensor $k$'s updates $A_k^{n-1}$, $Z_k^{n+1}$, and $q_k^{n+1}$ for the next transmission $n+1$.

\section{Federated Learning with Correlated Data}\label{Sec: non-iid FL}

\subsection{Federated GPD-Model Learning}
Assume that the sensor collects some historical data of the inter-arrival time $x$ to estimate the GPD model $\boldsymbol{\theta}$ before proceeding with problem \eqref{origin}.
Given the set $\mathcal{Y}_k=\{y:y|_{-\ln (x_k)>x_0}=-\ln (x_k)-x_0\},\forall\,k\in\mathcal{K}$, of the empirical data of exceedances, each sensor $k$ locally finds the GPD distribution which is the closest to the empirical distribution of $\mathcal{Y}_k$ in terms of the Kullback--Leibler (KL) divergence  $D(\mathcal{Y}_k\lvert\rvert\phi)=\sum_{y\in\mathcal{Y}_k}\frac{1}{\lvert\mathcal{Y}_k\rvert}\ln\big(\frac{1/\lvert\mathcal{Y}_k\rvert}{\phi(\boldsymbol{\theta} | y) {\rm d}y}\big)$. Here, $\phi(\boldsymbol{\theta}|y)= \frac{1}{\sigma}\big( 1+ \frac{\xi y}{\sigma}\big )^{-(1+1/\xi)}$ is the likelihood function. To this goal, we minimize the KL divergence as $\underset{\boldsymbol{\theta}}{\max}~\frac{1}{\lvert\mathcal{Y}_k\rvert}\sum_{y\in\mathcal{Y}_k}\ln\phi(\boldsymbol{\theta} | y)$ which can be solved via gradient ascent. That is, each sensor $k$ iteratively updates
\begin{equation}\label{Eq: Local gradient descent}
  \boldsymbol{\theta}^{j}_k=\boldsymbol{\theta}^{j-1}_k+ \frac{\gamma }{\lvert\mathcal{Y}_k\rvert} \sum_{y\in\mathcal{Y}_k} \nabla_{\boldsymbol{\theta}}
\ln\phi(\boldsymbol{\theta}^{j-1}_k | y)
\end{equation}
with the learning rate $\gamma$ and gradient
\begin{align*}
\nabla_{\boldsymbol{\theta}}\ln\phi(\boldsymbol{\theta} | y) =
\Bigg( \frac{\xi+1}{\frac{\sigma^2}{y}+\sigma \xi} -\frac{1}{\sigma},
 \frac{1}{\xi^2}\ln\Big(1+ \frac{\xi y}{\sigma}\Big)-\frac{1+\frac{1}{\xi}}{\frac{ \sigma}{y}+ \xi  }\Bigg).
\end{align*}
Additionally, we let all sensors have an identical initial value $\boldsymbol{\theta}^{0}$ in gradient ascent.
Note that $\mathcal{Y}_k$ is composed of the exceedance data for tail distribution characterization. Hence, given a moderate\footnote{If we consider the online GPD-model training for problem \eqref{origin}, the URLLC constraint \eqref{outg} cannot be addressed within this duration.} data-collecting time duration,  the sensor may not have enough data to achieve a sufficiently accurate estimation. Although a more accurate GPD model can be obtained by aggregating all sensors' local data at the central controller, uploading the local data incurs extra transmit power which is precious for the battery-limited sensor.
In order to diminish the overhead while preserving the controller's global view, we adopt the FL framework in which the sensors instead upload their locally-trained GPD models $ \boldsymbol{\theta}^{J}_k,\forall\,k\in\mathcal{K}$, after the convergence in \eqref{Eq: Local gradient descent} is achieved, e.g., the completion of $J$ iterations. Then the controller finds the global  GPD model $\boldsymbol{\theta}_{\rm GL}= \frac{\sum_{k\in\mathcal{K}}\lvert\mathcal{Y}_k\rvert \boldsymbol{\theta}^{J}_k}{\sum_{k\in\mathcal{K}}\lvert\mathcal{Y}_k\rvert}$ by weighted average \cite{MahMooRamHamArc17} and feeds it back to the sensors.

\subsection{Correlation-Aware Local-Model Selection}

The global model $\boldsymbol{\theta}_{\rm GL}$ and all local models $ \boldsymbol{\theta}^{J}_k,\forall\,k\in\mathcal{K}$, are stochastic due to the randomness of the empirical data in $\mathcal{Y}_k$. As a consequence,
the variance\footnote{For notational simplicity, $\mbox{Var}(\boldsymbol{\theta})$ represents the variance of $\sigma$ or $\xi$.} of the global model, i.e.,
\begin{equation}\label{Eq: Global model variance}
\mbox{Var}(\boldsymbol{\theta}_{\rm GL})= \frac{\sum_{k\in\mathcal{K}}\lvert\mathcal{Y}_k\rvert ^2\mbox{Var}(\boldsymbol{\theta}^{J}_k)}{(\sum_{k\in\mathcal{K}}\lvert\mathcal{Y}_k\rvert)^2},
\end{equation}
will affect the performance of power consumption, peak AoI, and queuing delay of the studied industrial IoT system. To deduce the details of the variance $\mbox{Var}(\boldsymbol{\theta}^{J}_k)$, let us intuitively express
\begin{equation}
\boldsymbol{\theta}^{J}_k=g\bigg(\frac{\gamma}{\lvert\mathcal{Y}_k\rvert}  \sum_{y\in\mathcal{Y}_k}\nabla_{\boldsymbol{\theta}}\ln\phi(\boldsymbol{\theta}^0 | y)\bigg)
\end{equation}
based on \eqref{Eq: Local gradient descent} with a function $g(\cdot)$. By further referring to \cite{BenHanNag05}
\begin{equation}
\mbox{Var} (g(X))\approx [g'(\mathbb{E} [X])]^{2}\mbox{Var} (X),
\end{equation}
we can derive
\begin{align}
&\mbox{Var}(\boldsymbol{\theta}^{J}_k)\approx\frac{\kappa\gamma^2}{\lvert\mathcal{Y}_k\rvert^2} \mbox{Var}\bigg(  \sum_{y\in\mathcal{Y}_k}\nabla_{\boldsymbol{\theta}}\ln\phi(\boldsymbol{\theta}^0 | y)\bigg)\label{Eq: Variance local}
\end{align}
with $\kappa= [g'(\gamma\mathbb{E} [\nabla_{\boldsymbol{\theta}}\ln\phi(\boldsymbol{\theta}^0 | y)])]^{2}$ and, moreover,
\begin{multline}
\mbox{Var}\bigg( \sum_{y\in\mathcal{Y}_k}\nabla_{\boldsymbol{\theta}}\ln\phi(\boldsymbol{\theta}^0 | y)\bigg)
=\lvert \mathcal{Y}_k\rvert \mbox{Var}\big(\nabla_{\boldsymbol{\theta}}\ln\phi(\boldsymbol{\theta} ^0| y)\big)
\\ + \sum_{y,\tilde{y}\in\mathcal{Y}_k|y\neq  \tilde{y}}  \mbox{Cov}(\nabla_{\boldsymbol{\theta}}\ln\phi(\boldsymbol{\theta}^0 | y),\nabla_{\boldsymbol{\theta}}\ln\phi(\boldsymbol{\theta}^0 | \tilde{y})\big).\label{Eq: Variance single}
\end{multline}
If the inter-arrival time is correlated as assumed in Section \ref{Sec: System}, the covariance will be larger than zero. A stronger correlation between the empirical data further increases the variance $\mbox{Var}(\boldsymbol{\theta}_{\rm GL})$. Motivated by this, we select (a part of sensors') local models for weighted average in FL by accounting for the data correlation. To this end, let us consider a discrete-time stochastic process $\{X_t\}$. The process is long-range dependent (LRD) if the normalized auto-covariance function decays hyperbolically in the asymptotic manner, i.e., $\mbox{K}_{XX}(m)/\mbox{Var}(X) \sim m^{-\alpha}$ with $0<\alpha<1$. The process is short-range dependent  (SRD) if the auto-covariance function decays exponentially or faster. The dependence feature is also reflected by the Hurst exponent $H$. For the LRD process, we have $0.5<H<1$ and $\alpha=2-2H$ \cite{BerSheTaqWil95}. The dependence is stronger as $H\to 1$. Additionally, the SRD process has $H=0.5$. The Hurst exponent can be found via the rescaled range (R/S) analysis \cite{BerSheTaqWil95}.  Applying $\mbox{K}_{XX}(m)\sim \mbox{Var}(X) m^{2H-2}$ to $\mbox{Cov}(\cdot,\cdot)$ in \eqref{Eq: Variance single} and incorporating \eqref{Eq: Global model variance}, \eqref{Eq: Variance local}, and \eqref{Eq: Variance single}, we derive
\begin{multline}\label{Eq: normalized variance}
\frac{\mbox{Var}(\boldsymbol{\theta}_{\rm GL})}{ \mbox{Var}(\nabla_{\boldsymbol{\theta}}\ln\phi(\boldsymbol{\theta}^0 | y))}
\\\leq \frac{\kappa\gamma^2\sum_{k\in\mathcal{K}}(\lvert\mathcal{Y}_k\rvert+2 \sum_{i=1}^{\lvert\mathcal{Y}_k\rvert} \sum_{m=1}^{\lvert\mathcal{Y}_k\rvert-i} m^{2H_k-2}) }{(\sum_{k\in\mathcal{K}}\lvert\mathcal{Y}_k\rvert)^2}
\end{multline}
in which the inequality is established since the exceedance data of  inter-arrival time are acquired intermittently.
Subsequently,  referring to \eqref{Eq: normalized variance}, we define a cost function
\begin{equation*}
\Psi(\boldsymbol{\eta})=\frac{\sum_{k\in\mathcal{K}}\eta_k(\lvert\mathcal{Y}_k\rvert+2 \sum_{i=1}^{\lvert\mathcal{Y}_k\rvert} \sum_{m=1}^{\lvert\mathcal{Y}_k\rvert-i} m^{2H_k-2}) }{(\sum_{k\in\mathcal{K}}\eta_k\lvert\mathcal{Y}_k\rvert)^2}
\end{equation*}
and focus on the variance minimization problem
 \begin{IEEEeqnarray}{cl}\label{Eq: Model selection}
 \underset{\eta_k\in\{0,1\}}{\mbox{minimize}}
 &~~\Psi(\boldsymbol{\eta})
 \end{IEEEeqnarray}
for selecting the local models.  $\boldsymbol{\eta}=(\eta_k:k\in\mathcal{K})$ is the model selection vector. In \eqref{Eq: Model selection}, we neglect $m^{2H_k-2}$ if sensor $k$ has the SRD data. Note that using the time-consuming exhaustive search to solve problem \eqref{Eq: Model selection} requires us to check all $2^K$ values of the objective. Alternatively, we invoke the notion of swap matching \cite{BodLeeChoHasWie11} in matching theory whose complexity is in the order of $\mathcal{O}(K^2)$ \cite{LiuBenDebPoo19}. Let us illustrate the swap matching-based method as follows. Firstly we are given a specific vector $\boldsymbol{\eta}$.  Additionally consider another vector $\tilde{\boldsymbol{\eta}}$ by either altering the value of a randomly-chosen element  $\eta_k$ in $\boldsymbol{\eta}$ or choosing a pair $(\eta_k,\eta_{k'})=(1,0)$ of $\boldsymbol{\eta}$ and swapping their values as $(\eta_k,\eta_{k'})=(0,1)$. If $\Psi(\tilde{\boldsymbol{\eta}}) <  \Psi(\boldsymbol{\eta})$, replace $\boldsymbol{\eta}$ with $\tilde{\boldsymbol{\eta}}$. We repeatedly check whether an alternative model selection vector $\tilde{\boldsymbol{\eta}}$ with the smaller cost $\Psi(\tilde{\boldsymbol{\eta}}) $ exists for the current $\boldsymbol{\eta}$. The steps of the proposed correlation-aware model selection approach for FL are outlined in Algorithm \ref{Alg: swap matching}. After finding the solution $\boldsymbol{\eta}^{*}$, the global model is calculated as $\boldsymbol{\theta}_{\rm GL}= \frac{\sum_{k\in\mathcal{K}} \eta^{*}_k \lvert\mathcal{Y}_k\rvert\boldsymbol{\theta}^{J}_k}{\sum_{k\in\mathcal{K}}  \eta^{*}_k \lvert\mathcal{Y}_k\rvert}$.
\begin{algorithm}[t]
\caption{Local-Model Selection for Federated Learning}
\begin{algorithmic}[1]
\State Initialize $\boldsymbol{\eta}$ and calculate the cost $ \Psi(\boldsymbol{\eta})$.
\Repeat 
\State  Based on $\boldsymbol{\eta}$, find an alternative $\tilde{\boldsymbol{\eta}}$  and calculate $ \Psi(\tilde{\boldsymbol{\eta}})$.
\If{$\Psi(\tilde{\boldsymbol{\eta}}) <  \Psi(\boldsymbol{\eta})$}
\State Update $ \boldsymbol{\eta} \gets \tilde{\boldsymbol{\eta}} $ and $\Psi(\boldsymbol{\eta}) \gets \Psi(\tilde{\boldsymbol{\eta}})$.
\EndIf
\Until{No alternative $\tilde{\boldsymbol{\eta}}$ with the smaller cost $ \Psi(\tilde{\boldsymbol{\eta}})$ exists.}
\end{algorithmic}
\label{Alg: swap matching}
\end{algorithm}
\begin{table}[t]
\caption{Simulation Parameters}
\centering
\begin{tabular}{|>{\centering}m{0.55cm}|>{\centering}m{0.65cm}||>{\centering}m{0.55cm}|>{\centering}m{0.8cm}||>{\centering}m{0.65cm}|>{\centering\arraybackslash}m{1.7cm}|}
\hline
 Para. &   Value & Para. & Value &  Para. &   Value
\\\hline
$K$   &   50   & $W$   & 1\,MHz    &   $P_{\max}$   &   10\,dBm 
\\\hline
$\beta$ &  1 & $D$   & 10\,kbit    &$N_0$   &  -174\,dBm/Hz
\\\hline
  $\gamma$  &   0.01   &  $\epsilon$  & $10^{-4}$     & $x_0$  & $\bar{F}^{-1}_X(0.01)$
\\\hline
$f_{\rm th}$  &  0.25 &   $J$   &   3000  &   $q_{\rm th}$   &   0.2\,sec
\\\hline
\end{tabular}\label{Tab: sparameter}
\end {table}

\section{Numerical Results}
In simulations, we consider the path loss model $33 \log x + 20\log 2.625 +32$ (dB) at the 2.625\,GHz carrier frequency \cite{rpt:itu_indoor} in which $x=15$\,m represents the distance between the sensor and controller. The inter-arrival time follows a folded normal distribution with the mean 0.1\,sec. The rest of the simulation parameters are listed in Table \ref{Tab: sparameter}.

\begin{figure}[t]
\includegraphics[width =0.92\columnwidth]{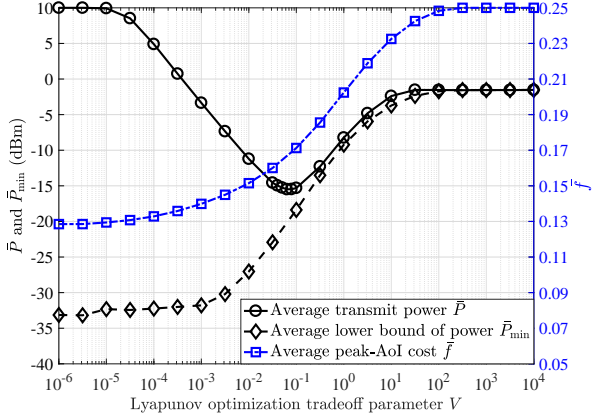}
\caption{Average transmit power, average lower bound of transmit power, and average peak-AoI cost versus $V$.}
\label{Fig: Power_AoI}
\end{figure} 
\begin{figure}[t]
\includegraphics[width =0.92 \columnwidth]{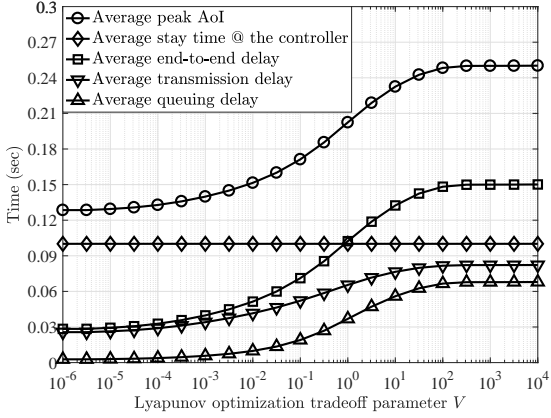}
\caption{Average peak AoI, average stay time of the status data at the controller, and average delays versus $V$.}
\label{Fig: AoI_delay}
\end{figure} 
In Figs.~\ref{Fig: Power_AoI} and \ref{Fig: AoI_delay}, we show the average performance of the sensor's transmit power, information age, and delays by varying the tradeoff parameter $V$.
It can be straightforwardly understood that raising $V$ decreases the sensor's transmit power at the expense of the higher information age as per problem \eqref{Eq: Per slot problem}. Note that the lower transmit power results in the higher transmission delay which consequently increases the queuing delay of the next status data. Accordingly, the average age cost  $\bar{f}$, peak AoI, end-to-end delay, transmission delay, and queuing delay monotonically increase with $V$.
Since the queuing delay is recursively related and affected by the transmission delay of the previous status data, lowering the transmit power (i.e., increasing $V$) has the higher impacts on the queuing delay in contrast with the transmission delay. 
Additionally, due to the higher power requirement \eqref{out transfer 3} of the status data with a higher queuing delay, the average lower bound $\bar{P}_{\min}$ of the sensor's transmit power increases with $V$ as shown in Fig.~\ref{Fig: Power_AoI}.
When the tradeoff parameter $V$ is larger than $10^{-1.2}$, constraint \eqref{out transfer 3} dominates in the power minimization problem \eqref{Eq: Per slot problem}, making  average transmit power $\bar{P}$  and $\bar{P}_{\min}$ almost coincide.  Owing to this rationale, the curve of $\bar{P}$ shows cavity. Thus, unlike most Lyapunov optimization-enabled resource allocation policies in which the optimal solutions are asymptotically obtained by letting $V\to\infty$, our optimal average power consumption is achieved at a finite $V$, i.e., $10^{-1.2}$, in the simulated setting.
Fig.~\ref{Fig: AoI_delay} also shows the average stay time of the status data at the controller which is is equal to the average inter-arrival time at the sensor. That is, the controller's data-updating frequency is identical to the sensor's data-sampling frequency.
\begin{figure}[t]
	\includegraphics[width = 0.92\columnwidth]{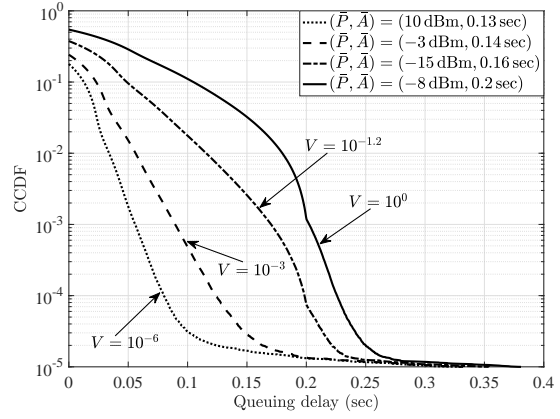}
	\caption{CCDF of the queuing delay for various $V$.}
	\label{Fig: Delay_tail}
\end{figure} 
\begin{figure}[t]
	\includegraphics[width =0.92 \columnwidth]{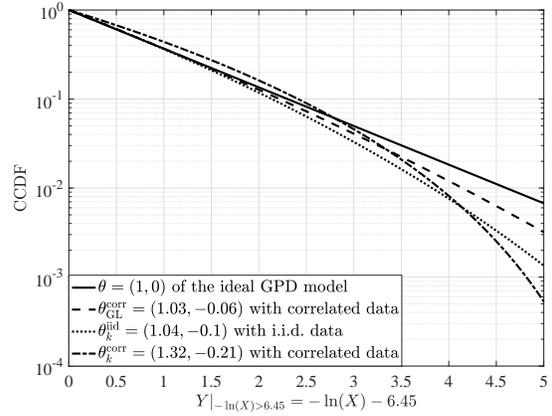}
	\caption{CCDFs of the ideal and estimated GPD models.}
	\label{Fig: Tail_accuracy}
\end{figure} 
Let us further investigate the CCDF of the queuing delay in Fig.~\ref{Fig: Delay_tail}. Therein, the smaller $\bar{A}$, i.e., average peak AoI, contains not only the lower average queuing delay but also the steeper decay in the tail distribution.
In contrast with the case $V=10^{-1.2}$, the probabilistic queuing delay constraint in the case  $V=10^{0}$ is not satisfied even though the average power consumption is higher. This is caused by the inefficient power utilization when $V> 10^{-1.2}$.

Subsequently, by considering that $K=10$ sensors have the status data with correlated inter-arrival time, Fig.~\ref{Fig: Tail_accuracy} shows the CCDFs of the ideal and estimated GPD models of the exceedances $Y|_{-\ln(X)>x_0}=-\ln(X)-x_0$. Therein, the ideal GPD parameters of the simulated setting are $\boldsymbol{\theta}=(1,0)$. All sensors have the identical correlation strength. For each sensor, the data amount of exceedances is $\lvert\mathcal{Y}_k\rvert=80$.
In Fig.~\ref{Fig: Tail_accuracy}, we consider the CCDF of the sensor's GPD model $\boldsymbol{\theta}_k^{\rm corr}$ which has the largest deviation in the distribution tail among all sensors. 
In contrast with the case $\boldsymbol{\theta}_k^{\rm iid}$ in which the inter-arrival time is independent and identically distributed (i.i.d.), the correlation has a higher impact on the estimation of the GPD parameters.  
As expected, the estimation accuracy with data correlation can be improved by leveraging the proposed approach for FL.  In this regard, the learned global GPD model $\boldsymbol{\theta}_{\rm GL}^{\rm corr}$ is closer to the ideal GPD model.
Finally,  we compare our proposed model selection approach with the correlation-agnostic baseline {\bf FedAvg} in Table \ref{Tab: FL performance} which shows the standard deviations of the learned global GPD models. 
Note that all sensors' correlation strengths are different, and the baseline has $\eta^{*}_k=1,\forall\,k\in\mathcal{K}$ \cite{MahMooRamHamArc17}.
For each sensor, the total data number of inter-arrival time (for finding the Hurst exponent by the R/S analysis) is approximately $100 \cdot \lvert\mathcal{Y}_k\rvert$ since we set $x_0=\bar{F}^{-1}_X(0.01)$. 
Verified by the results, our proposed local-model selection approach achieves a lower standard deviation of the global GPD model $\boldsymbol{\theta}_{\rm GL}$ when the sensor has less data samples of exceedances. In this regime, data correlation has a higher impact on the GPD-model learning.

\begin{table}[t]
\centering
\caption{Standard Deviation of $\boldsymbol{\theta}_{\rm GL}=(\sigma_{\rm GL},\xi_{\rm GL})$}
\begin{tabular}{|>{\centering}m{0.5cm}||>{\centering}m{2cm}|>{\centering\arraybackslash}m{2cm}|}
\hline
$\lvert\mathcal{Y}_k\rvert$  &  Proposed & {\bf FedAvg} 
\\\hline
\hline
 $5$  &$(0.4549,0.2951) $   &$(0.4817,0.3547) $
\\\hline
$10$  & $(0.2341,0.1428) $ &$(0.2579,0.1453) $
\\\hline
$20$ &$(0.1247,0.0832) $  &$(0.1254,0.0842) $
\\\hline
\end{tabular}
\label{Tab: FL performance}
\end {table}

\section{Conclusion}
In this work, we jointly took into account the peak AoI and delay distribution tail while allocating the sensor's transmit power by Lyapunov optimization. The studied problem was formulated  as a transmit power minimization in which the tail distribution is approximated as a GPD. We have further incorporated the FL framework for training the GPD model and proposed a correlation-aware local-model selection approach for FL. Finally, we have investigated the power-delay-AoI tradeoff and verified the effectiveness of our correlation-aware approach for FL.

\section*{Acknowledgments}
This research was supported by the Academy of Finland project MISSION,  the Academy of Finland project SMARTER, the CHIST-ERA project LeadingEdge under Grant CHIST-ERA-18-SDCDN-004,  the CHIST-ERA project CONNECT, the INFOTECH project NOOR, and the Nokia Bell Labs project NEGEIN.

%

\bibliographystyle{IEEEtran}
\bibliography{References}

\end{document}